**SODA: A Natural Language Processing Package to Extract Social Determinants of Health from Clinical Narratives**

Authors:
Zehao Yu, MS[1]
Xi Yang, PhD[1,2]
Chong Dang, MS[1]
Prakash Adekkanattu, PhD[3]
Braja Gopal Patra, PhD[4]
Yifan Peng, PhD[4]
Jyotishman Pathak, PhD[4]
Debbie L. Wilson, PhD, RN[5]
Ching-Yuan Chang, PhD[5]
Wei-Hsuan Lo-Ciganic, PhD[5]
Thomas J. George[6]
William R. Hogan, MD[1]
Yi Guo, PhD[1,2]
Jiang Bian, PhD[1,2]
Yonghui Wu, PhD[1,2]

Affiliation of the authors:
[1]Department of Health Outcomes and Biomedical Informatics, College of Medicine, University of Florida, Gainesville, Florida, USA
[2]Cancer Informatics Shared Resource, University of Florida Health Cancer Center, Gainesville, Florida, USA
[3]Information Technologies and Services, Weill Cornell Medicine, New York, NY, USA.
[4]Department of Population Health Sciences, Weill Cornell Medicine, New York, NY, USA.
[5]Department of Pharmaceutical Outcomes & Policy, College of Pharmacy, University of Florida, Gainesville, FL 32611, USA
[6]Division of Hematology & Oncology, Department of Medicine, College of Medicine, University of Florida, Gainesville, Florida, USA

Corresponding author:
Yonghui Wu, PhD
Clinical and Translational Research Building, 2004 Mowry Road, PO Box 100177, Gainesville, FL, USA, 32610
Phone: 352-294-8436, Email: yonghui.wu@ufl.edu

Word count:

**ABSTRACT**

**Objective**

This study aims to develop a natural language processing (NLP) package, SODA (i.e., SOcial DeterminAnts), to extract social determinants of health (SDoH) from clinical narratives, examine the generalizability of SODA for patients with different diseases (i.e., cancer and opioid use), and examine patient-level extraction rates for 19 SDoH categories.

**Methods**

We identified common SDoH categories and attributes and developed SDoH corpora using clinical notes from cancer and opioid patient cohorts. We systematically compared seven transformer-based large language models (LLMs) and developed an open-source package – SODA. We examined the generalizability of SODA using two disease domains including cancer and opioid use in clinical narratives from the University of Florida (UF) Health and explored strategies to improve performance. We applied SODA to extract 19 categories of SDoH from the breast (n=7,971), lung (n=11,804), and colorectal cancer (n=6,240) cohorts to assess patient-level extraction rates.

**Results**

We developed an SDoH corpus using 629 clinical notes of cancer patients with annotations of 13,193 SDoH concepts/attributes from 19 categories of SDoH, and another cross-disease validation corpus using 200 notes from opioid use patients with 4,342 SDoH concepts/attributes. We compared seven transformer models and the Bidirectional Encoder Representations from Transformers (BERT) model achieved the best strict/lenient F1 scores of 0.9147 and 0.9441 for SDoH concept extraction and 0.9617 and 0.9626 for linking attributes to SDoH concepts. The performance dropped when we applied the cancer SDoH model to the opioid cohort; fine-tuning using a smaller opioid SDoH corpus improved the strict/lenient F1 scores from 0.8172/0.8502 to 0.8312/0.8679. The extraction rates among varied in the three cancer cohorts, in which 10 SDoH could be extracted from over 70% of cancer patients, but 9 SDoH had extraction rates lower than 70%.

**Conclusions**

Our SODA package achieved good performance in extracting 19 categories of SDoH from clinical narratives of cancer and opioid use population identified at the UF Health. The SODA package with pre-trained transformer models is publicly available at https://github.com/uf-hobi-informatics-lab/SDoH_SODA.

**Statement of Significance**

**Problem**

There is no out-of-shelf NLP package for SDoH extraction from clinical narratives.

**What is Already Known**

Previous studies have applied NLP models to extract a limited number of SDoH categories, but there are no out-of-shelf packages available. Transformer-based models have been applied individually, but there is no systematic comparison. It's not clear how they perform when applied to different disease domains and the patient-level extraction rate is not clear.

**What This Paper Adds**

This study compared seven state-of-the-art transformer models for SDoH extraction, examined the generalizability across two disease domains including cancer and opioid use, and contributed an open-source package, SODA, to fill the gap of using SDoH in retrospective cohort studies.

# 1. INTRODUCTION

Social [e.g., education] and behavioral [e.g., smoking] determinants of health (hereafter SDoH for simplicity) are increasingly recognized as important factors affecting a wide range of health, functional, and quality of life outcomes, as well as healthcare fairness and disparities. For example, up to 75% of cancer occurrences are associated with SDoH, [1] which affect individual cancer risks and influence the likelihood of survival, early prevention, and health equity. [2–4] SDoH are associated with the frequency of opioid use and are important factors in preventing opioid misuse. [5–7] Various national and international organizations, such as the World Health Organization (WHO) [8], Healthy People 2030 [9], American Hospital Association (AHA) [10], National Institutes of Health (NIH), and Centers for Disease Control and Prevention (CDC) [11] have unanimously highlighted the importance of SDoH to people's health. There is an increasing interest in studying the role of SDoH in health outcomes and healthcare disparities, yet SDoH are not well-documented in electronic health records (EHRs). In February 2018, the International Classification of Diseases, Tenth Revision, Clinical Modification (ICD-10-CM) Official Guidelines for Coding and Reporting approved that healthcare providers involved in the care of a patient can document SDOH using Z codes (Z55–Z65); however, current reporting of SDoH using ICD-10-CM Z codes is relatively low (2.03% at patient-level) [12] and most individual-level SDoH are only documented in clinical narratives. [13] Natural language processing (NLP) systems that extract comprehensive SDoH information from clinical narratives are needed.

SDoH are often referred to as factors related to the conditions and status where people are born, live, and work, and are distinct from medical determinants of health (MDoH, e.g., diseases, medical procedures) from healthcare. [9] The definition of SDoH varies across different organizations. Still, common SDoH categories usually include economic stability, education access and quality, social and community context, neighborhood and built environment, and healthcare access and quality. [9] There is growing evidence on the significant association of SDoH with healthcare outcomes such as mortality [14], morbidity [15], mental health status [16], functional limitations [17], and substance use including opioid crisis [7]. For example, Galea *et al.* [18] estimated the number of cancer deaths attributable to SDoH in the United States and reported that low education, racial segregation, low social support, poverty, and income inequality attributed to cancer deaths were comparable to pathophysiological and behavioral causes. Albright *et al.* [5] identified education, housing stability, and employment status significantly associated with the frequency of opioid abuse.[5,6] Cantu *et al.* [7] examined three counties with opioid misuse in Ohio and identified social and economic instability such as unemployment, criminalization of substance use, limited access to healthcare, poverty, and social isolation among the root causes. As SDoH are not well-documented in structured EHRs, many studies [8,19,20] have explored SDoH collected using surveys.

Extracting SDoH from clinical narratives is a typical task of clinical concept extraction or named entity recognition (NER), which identifies phrases of interest (represented using the beginning position and ending position in the text) and determines their semantic categories (e.g., homelessness, smoking). Previous studies [13] have applied NLP methods to extract a single SDoH category from clinical narratives such as homelessness and housing insecurity [21,22], employment status [23], suicide detection [24], marital status [25], and substance use [26,27]. Rule-based and traditional machine learning models have been applied. Recent studies developed corpora with multiple common SDoH categories and applied deep learning-based NLP models. Yetisgen *et al.* [28] developed a corpus of 13 SDoH categories using notes from the publicly available MTSample dataset; Lybarger *et al.* [29] developed a corpus of 12 SDoH using clinical notes from the University of Washington and applied deep learning models including bidirectional long short-term memory (bi-LSTM) and BERT; Feller *et al.* [30] developed a corpus of 5 SDoH categories using notes from Columbia University Medical Center and applied traditional machine learning models; Stemerman *et al.* [16] developed a corpus of 6 SDoH categories and applied the BI-LSTM model; Gehrmann *et al.* [31] and Han *et al.* [32] explored SDoH using clinical notes from the Medical Information Mart for Intensive Care III (MIMIC-III) dataset; Feller *et al.* [33] developed a corpus of 6 SDOH categories using notes from Columbia University Irving Medical Center. We also have developed an SDoH corpus and transformer-based NLP methods [34], examined the extraction rate for a lung cancer cohort [35], and identified potential disparity for treatment options in a type 2 diabetes cohort [36].

Most recent studies for SDoH often applied deep learning models [37]. The 2022 n2c2 organized an NLP challenge focusing on SDoH, which greatly improved the adoption of transformer-based large language models (LLMs)[38]. Recent studies have explored transformer architectures such as BERT and RoBERTa [39,40]. Most NLP methods for SDoH were developed without a disease domain, yet researchers must apply these methods to a disease-specific cohort to study the role of SDoH in EHR-based retrospective cohorts. It is unclear how well current NLP systems can be used to extract SDoH for retrospective patient cohorts and across different disease domains. Until now, there is no off-the-shelf NLP package to facilitate the use of SDoH for EHR-based studies.

The goals of this study were (1) to develop an SDoH corpus and an open-source NLP package, SODA (i.e., SOcial DeterminAnts), with pre-trained state-of-the-art transformer models for SDoH of cancer patients, (2) examine the generalizability of SDoH extraction across two disease domains including cancer and opioid use and explore strategies for customization, and (3) examine extraction rates for various SDoH categories in 3 cancer-specific (breast, lung, colorectal) cohorts. First, we developed a larger SDoH corpus using clinical notes of cancer and a smaller cross-disease validation corpus using opioid use patients identified at the University of Florida (UF) Health and compared transformer models including Bidirectional Encoder Representations from Transformers (BERT) [41] and RoBERTa [42], DeBERTa[43], Longformer[44], and GatorTron[45]. Then, we explored strategies to customize the cancer-specific NLP model to an opioid user cohort. Finally, we integrated SODA with pre-trained clinical models into an open-source software package.

## 2. METHODS

### 2.1 Data

This study used clinical narratives from the University of Florida (UF) Health Integrated Data Repository (IDR). The UF Health IDR is a clinical data warehouse that aggregates data from the university's various clinical and administrative information systems, including the Epic (Epic Systems Corporation) system. This study was approved by the UF Institutional Review Board (IRB #IRB201902362).

**General cancer cohort**: We identified a general cancer cohort between 2012 and 2020 in UF Health IDR using ICD-9 and ICD-10 cancer diagnoses codes, and randomly selected 20,000 cancer patients using stratified random sampling (by cancer types). Using this general cancer cohort, we identified and collected a total number of ~1.5 million clinical notes.

**Opioid use cohort**: We identified an opioid use cohort between 2016 and 2020 in UF Health IDR. Adult patients aged ≥18 who had at least one outpatient visit and at least one eligible opioid prescribing order (excluding injectable and buprenorphine approved for opioid use disorder). We excluded patients who had non-malignant cancers and who had their first opioid prescription order after Oct 1, 2019.

**Identify SDoH keywords:** We created a list of keywords to identify clinical notes that contained SDoH using a snowball strategy. We first collected seed keywords indicating SDoH from domain experts (TJG, WRH), healthcare representatives in stakeholders' panel meetings, as well as the biomedical literature. Then, we iteratively reviewed notes to identify new SDoH keywords and extend the seed SDoH keywords until there were no new keywords identified. A total of 30 SDoH keywords were identified.

**Training and test datasets from the cancer cohort**: We identified clinical notes containing SDoH by searching the 30 SDoH keywords in the general cancer cohort's clinical notes. Then, we identified clinical notes with at least three unique mentions of SDoH keywords and randomly sampled a subset for annotation. After annotation, we divided the annotated notes into a training set and a testing set with an 8:2 ratio and held out 10% of the training sample which we used as a validation set.

**SDoH annotation**: We reviewed SDoH categories defined by healthcare organizations and national agencies including the WHO, Healthy People 2030, and CDC and identified 7 main SDoH classes and 38 subclasses (**Figure 1**). We identified attributes for the 38 SDoH subclasses and developed initial annotation guidelines according to the SDoH definitions from different resources and iteratively fine-tuned the guidelines in training sessions. During the training sessions, the study team met routinely to identify and review the discrepancies in annotation. Our domain experts served as judges when the two annotators could not reach an agreement. We monitored the annotation agreement using Cohen's Kappa. When a good agreement score (>0.8) was achieved, the two annotators started annotation independently.

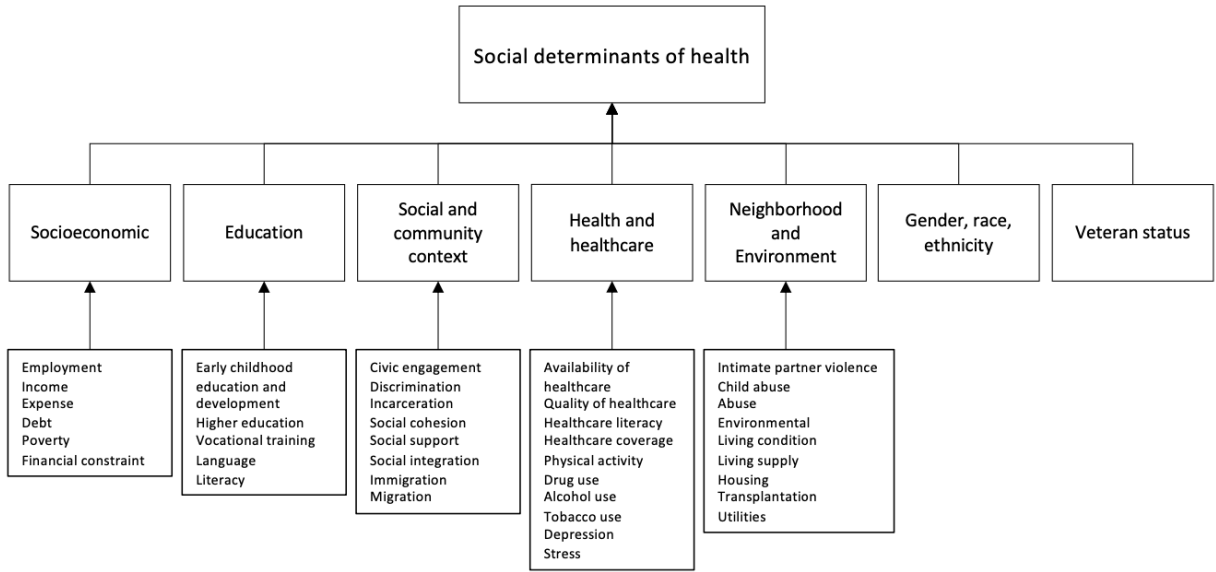

**Figure 1.** An overview of the 7 social determinants of health classes and 38 subclasses.

**Cross-disease evaluation dataset from an opioid use cohort**: We sought to examine how well the SDoH NLP models developed using cancer patients performed in a different cohort representing opioid use. We adopted the same procedure to identify clinical notes with at least three mentions of unique SDoH from the opioid use cohort and sampled a subset for annotation following the same annotation guidelines. We excluded cancer patients when sampling opioid notes for annotation to avoid any overlap between the two SDoH corpora. After annotation, we split the annotated notes into an additional fine-tuning set – used to fine-tune the cancer SDoH model, and a test set – used to evaluate the cross-disease performance on the opioid population.

**2.2 NLP methods to extract SDoH**

We approached SDoH extraction as a two-stage NLP task, including (1) a concept extraction step to identify SDoH concepts and attributes and (2) a relation extraction step to link the attribute to the targeted SDoH concepts. For example, "attend religious service" is a concept for "social cohesion" where "1 to 4 times per year" is an attribute indicating the frequency of attending religious service; "every day smoker" is an SDoH concept for "tobacco use" where "cigarettes", "1 packs/day", and "46 years" are the attributes indicating the smoking type, pack per day, and years of smoking, respectively. We explored pre-trained models from two state-of-the-art transformer architectures, BERT and RoBERTa. Our previous study showed that BERT and RoBERTa consistently outperformed other transformer models for clinical concept extraction [46]. Following our previous studies on clinical transformers, we examined pre-trained transformers from general English corpus (denoted as '_general', e.g., 'BERT_general') and clinical transformers pre-trained using clinical notes from the MIMIC-III database (denoted as '_mimic', e.g., 'BERT_mimic'). We adopted the default parameters optimized in our clinical transformer package [46]. We also explored new transformer architectures including DeBERTa [43], GatorTron [45], and Longformer [44].

**2.3 Identification of SDoH concepts and attributes using concept extraction**

We approached clinical concept extraction as a sequence labeling problem and adopted the 'BIO' labeling schema, where 'B-' and 'I-' are label prefixes indicating words at the beginning and inside of a concept, and 'O' stands for words located outside of any concepts of interests. We solved the task as a classification – for each word in a sentence, we determined a label in ['B', 'I', 'O']. In this study, we used the pre-trained transformer models to generate distributed word-level and sentence-level representations, then added a classification layer with Softmax activation to calculate a probability for each category. The cross-entropy loss was used for fine-tuning.

## 2.4 Linking attributes to core SDoH concepts using relation classification

The goal was to link attribute concepts (e.g., smoking frequency) to the core SdoH concept (e.g., tobacco use). Following our previous experience in relation classification, we approached attribute linking as a classification task – we generated candidate pairs of concepts and trained machine learning classifiers to classify them into predefined relation classes. We adopted a heuristic method developed in our previous studies [47,48] to identify candidate pairs of clinical concepts. Then, pre-trained transformer models were used to generate a distributed representation. To determine the relation type, we concatenated the contextual representations of the model special [CLS] token and all four entity markers and added a classification layer (a linear layer with Softmax activation) to calculate a score for each relation category. The cross-entropy loss was used in fine-tuning.

## 2.5 Evaluation and experiments design

**Evaluation methods:** We first evaluated SODA using a standard setting where both the training and test data were from a cancer cohort. We evaluated SODA on three subtasks including (1) a concept extraction task to extract SDoH concepts and attributes, (2) a relation extraction task to link attributes to the target SDoH concept (given ground-truth SDoH concepts), (3) an end-to-end task to extract SDoH concepts and link attributes to SDoH concepts. Then, we conducted a cross-disease evaluation to evaluate the NLP models using clinical notes from an opioid use cohort. We compared three application scenarios to evaluate SODA in cross-disease settings including (1) directly applying the NLP models developed for cancer patients to patients of opioid use, (2) merging the cancer corpus with the opioid fine-tuning corpus and training a model from scratch, and (3) fine-tuning the cancer SDoH model using the opioid fine-tuning set.

**Evaluation metrics**: Cohen's Kappa: We evaluated annotator agreement using Cohen's Kappa, κ, coefficient, where higher κ denotes better annotator agreement. We used the strict micro-averaged precision, recall, and F1-score aggregated from all classes to evaluate the concept extraction and relation extraction. The official evaluation scripts provided by the 2018 n2c2 challenge [49] were used to calculate these scores.

**Experimental setup:** We used pretrained transformer models developed in our previous study [46], where the transformer architecture was implemented using the Transformers library developed by the HuggingFace team in PyTorch. We fine-tuned transformer models using the training set. The best model was selected according to the validation performance measured by strict F1-scores on the validation set. We adopted an early stop strategy to stop the training when no improvements were observed in 5 consecutive epochs. We conducted all experiments using two Nvidia A100 GPUs.

## 3. RESULTS

We identified a total of 225,441 clinical notes containing at least three unique SDoH mentions from cancer patients and randomly sampled 700 for annotation. After de-duplicating and removing notes without valid SDoH annotations, there remained 629 notes in the cancer SDoH corpus. Two annotators (ZY and CD) annotated a total of 13,193 SDoH concepts in these notes. Among the 38 SDoH subclasses identified from various resources, there were 19 subclasses identified from the annotation. Table S1 (in the Supplement) provides the attributes identified for the 19 subclasses of SDoH. The inter-annotator agreement between the two annotators calculated by kappa score (using 20 overlapped notes) was low at 0.47 in the first training session, which was improved to 0.68 in the second round and eventually reached 0.89 after 5 iterative rounds of training followed by meetings to discuss and solve discrepancies. **Table 1** shows detailed numbers of concepts annotated for each SDoH category. From the opioid cohort, we identified ~13 million clinical notes from 98,074 patients. We followed the same annotation guidelines and annotated an SDoH corpus of 200 notes for cross-disease evaluation. **Table 2** shows the distribution of notes and SDoH concepts for training, validation, and test set of the two disease domains.

**Table 1.** Annotation results for the social determinants of health corpus from the cancer cohort.

| SDoH Class | Number of concepts | SDoH Subclasses | Number of concepts |
|---|---|---|---|
| Economic Stability | 596 | Financial constraint | 97 |
| | | Employment | 499 |
| Education | 602 | Language | 25 |
| | | Education | 577 |
| Health and Health care | 4,370 | Physical activity | 223 |
| | | SDoH ICD | 61 |
| | | Sexual activity | 637 |
| | | Drug use | 577 |
| | | Tobacco use | 1,998 |
| | | Alcohol use | 874 |
| Social and community context | 908 | Marital status | 488 |
| | | Social cohesion | 420 |
| Neighborhood and physical environment | 1,257 | Abuse (physical or mental) | 412 |
| | | Transportation | 193 |
| | | Living supply | 523 |
| | | Living condition | 129 |
| Gender, Race, and Ethnicity | 990 | Gender | 846 |
| | | Race | 110 |
| | | Ethnicity | 34 |

**Table 2.** Distribution of notes and SDoH in training, validation/fine-tuning, and test sets of the cancer cohort; and the opioid cohort.

| Disease domain | | Total # | Training/Fine-tuning | Validation | Test |
|---|---|---|---|---|---|
| Cancer | Total notes | 629 | 452 | 51 | 126 |
| | Total entities | 13,193 | 9,497 | 1,009 | 2,687 |
| | Entity/note | 20 | 21 | 20 | 21 |
| Opioid | Total notes | 200 | 90 | 10 | 100 |
| | Total entities | 4,342 | 1,952 | 173 | 2,217 |
| | Entity/note | 21 | 22 | 17 | 22 |

Table 3 compares seven transformer-based NLP models on three tasks including SDoH concept/attributes extraction, attribute linking, and end-to-end extraction (i.e., including both concept extraction and attribute linking). Strict and lenient scores were reported. For SDoH concept extraction, the BERT_general model trained using general English corpus achieved the best F1 strict/lenient scores of 0.9147 and 0.9441, respectively. GatorTron achieved the same strict F1 score as a tie. Table S2 (in the Supplement) provides detailed scores for each SDoH subclass. For attribute linking using relation classification, the BERT_general again achieved the best strict/lenient scores of 0.9617 and 0.9626, respectively. The end-to-end system using the BERT_general model achieved the best strict/lenient F1-scores of 0.8882 and 0.9146, respectively.

**Table 3.** Comparison of transformer models to identify SDoH concepts and link attributes on the cancer cohort.

| Task | Model | Strict | | | Lenient | | |
|---|---|---|---|---|---|---|---|
| | | Prec. | Rec. | F(b=1) | Prec. | Rec. | F(b=1) |
| Concept extraction to identify SDoH concepts and attributes | BERT_general | **0.9143** | 0.9151 | **0.9147** | **0.9436** | 0.9385 | **0.9411** |
| | BERT_mimic | 0.8870 | **0.9264** | 0.9063 | 0.9138 | **0.9507** | 0.9319 |
| | Roberta_general | 0.9061 | 0.9061 | 0.9061 | 0.9335 | 0.9311 | 0.9323 |
| | Roberta_mimic | 0.8987 | 0.9184 | 0.9084 | 0.9251 | 0.9437 | 0.9343 |
| | DeBERTa | 0.9025 | 0.9133 | 0.9079 | 0.9263 | 0.9346 | 0.9305 |

|  | | Prec. | Rec. | F1 | Prec. | Rec. | F1 |
|---|---|---|---|---|---|---|---|
| | GatorTron | 0.9115 | 0.918 | **0.9147** | 0.9338 | 0.9389 | 0.9364 |
| | Longformer | 0.8880 | 0.9182 | 0.9028 | 0.9176 | 0.9447 | 0.931 |
| Relation classification to link attributes to core SDoH concepts | BERT_general | 0.9584 | **0.9649** | 0.9617 | 0.9594 | **0.9659** | **0.9626** |
| | BERT_mimic | 0.9500 | 0.9630 | 0.9565 | 0.9510 | 0.9640 | 0.9574 |
| | Roberta_general | 0.9562 | 0.9348 | 0.9453 | 0.9572 | 0.9357 | 0.9463 |
| | Roberta_mimic | **0.9592** | 0.9387 | 0.9488 | 0.9602 | 0.9396 | 0.9498 |
| | DeBERTa | 0.9640 | 0.9396 | 0.9517 | **0.9650** | 0.9406 | 0.9527 |
| | GatorTron | 0.9572 | 0.9591 | 0.9582 | 0.9582 | 0.9601 | 0.9591 |
| | Longformer | 0.9612 | 0.9416 | 0.9513 | 0.9622 | 0.9426 | 0.9523 |
| End-to-end | BERT_general | **0.9134** | **0.8642** | **0.8882** | **0.9418** | **0.8890** | **0.9146** |
| | Gatortron | 0.9026 | 0.8588 | 0.8802 | 0.9219 | 0.8772 | 0.899 |

Best precision, recall, and F1-score are highlighted in bold. The official evaluation script developed by the 2018 n2c2 challenge were used to calculate the evaluation scores.

**Table 4.** Cross-disease evaluation results on the opioid test data set.

|  | Strict | | | Lenient | | |
|---|---|---|---|---|---|---|
|  | **Prec.** | **Rec.** | **F(b=1)** | **Prec.** | **Rec.** | **F(b=1)** |
| Direct evaluation | **0.8233** | 0.8111 | 0.8172 | **0.859** | 0.8417 | 0.8502 |
| Fine-tuning | 0.8186 | **0.8441** | **0.8312** | 0.8579 | 0.878 | 0.8679 |
| Merge and retrain | 0.8142 | 0.8427 | 0.8282 | 0.8572 | **0.8814** | **0.8691** |

Direct evaluation: directly evaluating the cancer SDoH model using opioid test set; Fine-tuning: fine-tuning the cancer model using the Opioid fine-tuning set; Merge and retrain: merging the Cancer training set and opioid fine-tuning set and retraining the model.

**Table 4** shows the results of the cross-disease evaluation. When directly applying BERT_general trained using cancer data to the opioid cohort, we observed a performance drop from strict/lenient F1 scores of 0.9216 and 0.9441 to 0.8172 and 0.8502, respectively. Both customization strategies improved the F1-score of SDoH extraction for opioid use patients. The best strict F1 score of 0.8312 was achieved by fine-tuning the cancer SDoH model using the opioid fine-tuning data.

Table 5 reports for three cancer cohorts the total number of SDoH concepts and the population-level extraction rate – defined as the total number of patients with at least one specific SDoH category divided by the total number of patients. For lung cancer, we identified a total of 11,804 patients with 1,796,131 notes. For breast cancer, we identified 7,971 patients with 1,143,304 clinical notes. For colorectal cancer, we identified 6,240 patients with 1,021,405 clinical notes. We applied the end-to-end NLP model to extract 19 SDoH categories and aggregated the SDoH to the patient level to examine the extraction rate.

Table 5. Number of SDoH instances and population-level extraction rate from lung, breast, and colorectal cancers.

| SDoH | Breast cancer # Concepts | Breast cancer Rate | Colorectal cancer # Concepts | Colorectal cancer Rate | Lung cancer # Concepts | Lung cancer Rate |
|---|---|---|---|---|---|---|
| Abuse (physical or mental) | 3,077 | 0.4674 | 1,378 | 0.3647 | 4,145 | 0.4284 |
| Alcohol use | 6,179 | 0.9387 | 3,598 | 0.9523 | 9,195 | 0.9503 |
| Drug use | 6,055 | 0.9199 | 3,521 | 0.9319 | 8,756 | 0.9050 |
| Education | 5,825 | 0.8849 | 3,370 | 0.8920 | 8,463 | 0.8747 |
| Ethnicity | 5,173 | 0.7859 | 2,509 | 0.6641 | 5,231 | 0.5406 |
| Financial constraint | 2,485 | 0.3775 | 981 | 0.2596 | 2,766 | 0.2858 |
| Gender | 6,486 | 0.9854 | 3,731 | 0.9875 | 9,552 | 0.9872 |
| Language | 5,158 | 0.7836 | 2,466 | 0.6527 | 5,173 | 0.5346 |
| Living condition | 3,192 | 0.4849 | 1,866 | 0.4939 | 5,359 | 0.5539 |
| Living supply | 5,853 | 0.8892 | 3,285 | 0.8695 | 7,861 | 0.8125 |
| Marital status | 6,015 | 0.9138 | 3,472 | 0.9190 | 8,655 | 0.8945 |
| Occupation/Employment | 5,882 | 0.8936 | 3,324 | 0.8798 | 8,345 | 0.8625 |
| Physical activity | 2,992 | 0.4545 | 1,136 | 0.3006 | 3,092 | 0.3195 |
| Race | 5,709 | 0.8673 | 3,087 | 0.8170 | 7,376 | 0.7623 |
| SDoH ICD | 562 | 0.0853 | 345 | 0.0913 | 1,239 | 0.1280 |
| Sexual activity | 5,606 | 0.8517 | 3,173 | 0.8398 | 8,124 | 0.8396 |
| Social cohesion | 2,458 | 0.3734 | 981 | 0.2596 | 2,727 | 0.2818 |
| Tobacco use | 4,940 | 0.7505 | 2,669 | 0.7064 | 7,639 | 0.7895 |
| Transportation | 2,524 | 0.3834 | 1,018 | 0.2694 | 2,877 | 0.2973 |

SDoH: social determinants of health; ICD: International Classification of Diseases; Rate: population-level extraction rate.

## 4. DISCUSSION AND CONCLUSION

NLP is the key technology to extract SDoH from clinical narratives. This study examined transformer-based NLP models for SDoH extraction from clinical narratives. We developed SDoH corpora from two disease domains (cancer and opioid use patients) with 19 SDoH categories and compared seven transformer-based NLP models for extraction. The end-to-end NLP system using the BERT-based transformer model achieved the best lenient F1-score of 0.9146. Our previous studies [46,48] showed that BERT_mimic outperformed BERT_general on extracting clinical concepts. This study showed that BERT_general (trained using general English corpora) outperformed BERT_mimic (fine-tuned using clinical text) for SDoH extraction. One potential reason is that most SDoH concepts are composed of general English words other than medical words. This could be the potential reason that there is limited benefit from GatorTron [45], an LLM trained using a larger clinical corpus.

In addition to the standard training/test evaluation, we conducted a cross disease cohort evaluation to examine how the cancer SDoH models perform on an opioid use cohort. We observed a performance drop when directly applying the cancer SDoH models to opioid use patients, indicating that the documentation of SDoH varied among different disease domains. We explored two strategies to customize the NLP model and the fine-tuning strategy achieved the best strict F1 score. The experimental results from the cross-disease evaluation showed that it is necessary to fine-tune the NLP module by annotating corpora from a new disease domain. We also examined the patient-level extraction rates for the 19 SDoH categories. The patient-level extraction rates were largely consistent among three cancer cohorts with some variations. For example, the lung cancer cohort had a higher extraction rate for tobacco use than the breast and colorectal cancer cohorts. This result is expected given the association between smoking and lung cancer and the strong emphasis on smoking cessation as a component of lung cancer therapy. There are 10 categories of SDoH extracted from > 70% population of the cancer patients, including gender, race, tobacco use, alcohol use, drug use, education, living supply, marital status, occupation, and sexual activity; 9 other categories had a relatively low extraction rate (< 70% population), indicating a potential gap of documenting SDoH in EHRs. In a previous study [50], we conducted a focused interview of stakeholders including oncologists, data analysts, citizen scientists, and patient navigators, and identified potential challenges and barriers to the low documentation rate of SDoH in EHRs, including lack of integration into clinical workflow, lack of incentives for SDoH data collection, and lack of training and tools for clinicians to derive actionable insights for decision making. Future studies should explore strategies to reduce these barriers and improve the documentation of SDoH in EHRs.

This study has limitations. First, a limited number of instances were annotated for some SDoH categories (e.g., language) due to low documentation rate. We plan to annotate more notes to increase the sample size. Similarly, the cross-disease performance of the NLP models could be further improved by annotating more opioid notes. Second, we may miss some keywords in the snowball procedure used to identify the seed SDoH keywords. Finally, the NLP models were developed using notes from patients with cancer and opioid use. Customization through fine-tuning is needed when applying SODA to other disease domains. Our future work will investigate how person-level SDoH affect cancer risks, treatment outcomes, and disparities.

ACKNOWLEDGMENTS

We gratefully acknowledge the support of NVIDIA Corporation and NVIDIA AI Technology Center with the donation of the GPUs and the computing resources used for this research. We acknowledge the support from the Cancer Informatics Shared Resource in the UF Health Cancer Center. The content is solely the responsibility of the authors and does not necessarily represent the official views of the funding institutions.

FUNDING STATEMENT

This study was partially supported by grants from the Patient-Centered Outcomes Research Institute® (PCORI®) (ME-2018C3-14754), the National Institute on Aging (1R56AG069880, 1R01AG080624-01, R21AG068717), the National Cancer Institute (1R01CA246418, 3R01CA246418-02S1, 1R21CA245858-01A1, R21CA245858-01A1S1, R21CA253394-01A1, R21CA253394-01A1), National Institute on Drug Abuse (1R01DA050676), National Institute of Mental Health (1R01MH121907, 5R21MH129682-02), National Library of Medicine (4R00LM013001), NSF Career (2145640), and Centers for Disease Control and Prevention (1U18DP006512).

COMPETING INTERESTS STATEMENT

The authors have no conflicts of interest that are directly relevant to the content of this study.

CONTRIBUTORSHIP STATEMENT

ZY, XY, JB, and YW were responsible for the overall design, development, and evaluation of this study. ZY and CD annotated the SDoH corpus from cancer patients' notes. DLW, CYC, and WL annotated the SDoH corpus from opioid use patients. TJG, WRH served as domain expert created seed keywords for SDoH and solved the discrepancies in the annotation. YG performed power calculations to determine the number of notes to annotate. PA, BGP, YP, and JP participated in the development of annotation guidelines. All authors reviewed the manuscript critically for scientific content, and all authors gave final approval of the manuscript for publication.

**REFERENCES**

1 Akushevich I, Kravchenko J, Akushevich L, *et al.* Cancer Risk and Behavioral Factors, Comorbidities, and Functional Status in the US Elderly Population. *ISRN Oncol* 2011;**2011**.https://www.ncbi.nlm.nih.gov/pmc/articles/PMC3197174/ (accessed 3 Feb 2019).
2 Hiatt RA, Breen N. The social determinants of cancer: a challenge for transdisciplinary science. *Am J Prev Med* 2008;**35**:S141-150.
3 Matthews AK, Breen E, Kittiteerasack P. Social Determinants of LGBT Cancer Health Inequities. *Semin Oncol Nurs* 2018;**34**:12–20.
4 Gerend MA, Pai M. Social determinants of Black-White disparities in breast cancer mortality: a review. *Cancer Epidemiol Biomarkers Prev* 2008;**17**:2913–23.
5 Albright DL, Johnson K, Laha-Walsh K, *et al.* Social Determinants of Opioid Use among Patients in Rural Primary Care Settings. *Soc Work Public Health* 2021;**36**:723–31.
6 Rangachari P, Govindarajan A, Mehta R, *et al.* The relationship between Social Determinants of Health (SDoH) and death from cardiovascular disease or opioid use in counties across the United States (2009–2018). *BMC Public Health* 2022;**22**:236.
7 Cantu R, Fields-Johnson D, Savannah S. Applying a Social Determinants of Health Approach to the Opioid Epidemic. *Health Promotion Practice* 2023;**24**:16–9.
8 Singh GK, Daus GP, Allender M, *et al.* Social Determinants of Health in the United States: Addressing Major Health Inequality Trends for the Nation, 1935-2016. *Int J MCH AIDS* 2017;**6**:139–64.
9 Social Determinants of Health - Healthy People 2030 | health.gov. https://health.gov/healthypeople/objectives-and-data/social-determinants-health (accessed 14 Sep 2021).
10 American Hospital Association| ICD-10-CM Coding for Social Determinants of Health. https://www.aha.org/system/files/2018-04/value-initiative-icd-10-code-social-determinants-of-health.pdf (accessed 2 Dec 2022).
11 Hillemeier M, Lynch J, Harper S, *et al.* Data Set Directory of Social Determinants of Health at the Local Level. ;:75.
12 Guo Y, Chen Z, Xu K, *et al.* International Classification of Diseases, Tenth Revision, Clinical Modification social determinants of health codes are poorly used in electronic health records. *Medicine (Baltimore)* 2020;**99**:e23818.
13 Patra BG, Sharma MM, Vekaria V, *et al.* Extracting social determinants of health from electronic health records using natural language processing: a systematic review. *J Am Med Inform Assoc* 2021;**28**:2716–27.
14 Sterling MR, Ringel JB, Pinheiro LC, *et al.* Social Determinants of Health and 90-Day Mortality After Hospitalization for Heart Failure in the REGARDS Study. *J Am Heart Assoc* 2020;**9**:e014836.

15 Eppes C, Salahuddin M, Ramsey PS, *et al.* Social Determinants of Health and Severe Maternal Morbidity During Delivery Hospitalizations in Texas [36L]. *Obstetrics & Gynecology* 2020;**135**:133S.
16 Stemerman R, Arguello J, Brice J, *et al.* Identification of social determinants of health using multi-label classification of electronic health record clinical notes. *JAMIA Open* Published Online First: 9 February 2021.https://doi.org/10.1093/jamiaopen/ooaa069 (accessed 7 Mar 2021).
17 May 10 EHP, 2018. Beyond Health Care: The Role of Social Determinants in Promoting Health and Health Equity. KFF. 2018.https://www.kff.org/racial-equity-and-health-policy/issue-brief/beyond-health-care-the-role-of-social-determinants-in-promoting-health-and-health-equity/ (accessed 12 Nov 2021).
18 Galea S, Tracy M, Hoggatt KJ, *et al.* Estimated Deaths Attributable to Social Factors in the United States. *Am J Public Health* 2011;**101**:1456–65.
19 Braveman P, Gottlieb L. The social determinants of health: it's time to consider the causes of the causes. *Public Health Rep* 2014;**129 Suppl 2**:19–31.
20 Chen X, Jacques-Tiura AJ. Smoking Initiation Associated With Specific Periods in the Life Course From Birth to Young Adulthood: Data From the National Longitudinal Survey of Youth 1997. *Am J Public Health* 2014;**104**:e119–26.
21 Gundlapalli AV, Carter ME, Palmer M, *et al.* Using Natural Language Processing on the Free Text of Clinical Documents to Screen for Evidence of Homelessness Among US Veterans. *AMIA Annu Symp Proc* 2013;**2013**:537–46.
22 Hatef E, Rouhizadeh M, Nau C, *et al.* Development and assessment of a natural language processing model to identify residential instability in electronic health records' unstructured data: a comparison of 3 integrated healthcare delivery systems. *JAMIA Open* 2022;**5**:ooac006.
23 Dillahunt-Aspillaga C, Finch D, Massengale J, *et al.* Using Information from the Electronic Health Record to Improve Measurement of Unemployment in Service Members and Veterans with mTBI and Post-Deployment Stress. *PLOS ONE* 2014;**9**:e115873.
24 Carson NJ, Mullin B, Sanchez MJ, *et al.* Identification of suicidal behavior among psychiatrically hospitalized adolescents using natural language processing and machine learning of electronic health records. *PLOS ONE* 2019;**14**:e0211116.
25 Bucher BT, Shi J, Pettit RJ, *et al.* Determination of Marital Status of Patients from Structured and Unstructured Electronic Healthcare Data. *AMIA Annu Symp Proc* 2020;**2019**:267–74.
26 Wang Y, Chen ES, Pakhomov S, *et al.* Automated Extraction of Substance Use Information from Clinical Texts. *AMIA Annu Symp Proc* 2015;**2015**:2121–30.
27 Rajendran S, Topaloglu U. Extracting Smoking Status from Electronic Health Records Using NLP and Deep Learning. *AMIA Jt Summits Transl Sci Proc* 2020;**2020**:507–16.
28 Yetisgen M, Vanderwende L. Automatic Identification of Substance Abuse from Social History in Clinical Text. In: ten Teije A, Popow C, Holmes JH, *et al.*, eds. *Artificial Intelligence in Medicine*. Cham: : Springer International Publishing 2017. 171–81.
29 Lybarger K, Ostendorf M, Yetisgen M. Annotating social determinants of health using active learning, and characterizing determinants using neural event extraction. *Journal of Biomedical Informatics* 2021;**113**:103631.
30 Feller DJ, Bear Don't Walk Iv OJ, Zucker J, *et al.* Detecting Social and Behavioral Determinants of Health with Structured and Free-Text Clinical Data. *Appl Clin Inform* 2020;**11**:172–81.

31  Gehrmann S, Dernoncourt F, Li Y, *et al.* Comparing deep learning and concept extraction based methods for patient phenotyping from clinical narratives. *PLOS ONE* 2018;**13**:e0192360.
32  Han S, Zhang RF, Shi L, *et al.* Classifying social determinants of health from unstructured electronic health records using deep learning-based natural language processing. *J Biomed Inform* 2022;**127**:103984.
33  Feller DJ, Zucker J, Don't Walk OB, *et al.* Towards the Inference of Social and Behavioral Determinants of Sexual Health: Development of a Gold-Standard Corpus with Semi-Supervised Learning. *AMIA Annu Symp Proc* 2018;**2018**:422–9.
34  Yu Z, Yang X, Dang C, *et al.* A Study of Social and Behavioral Determinants of Health in Lung Cancer Patients Using Transformers-based Natural Language Processing Models. *arXiv:210804949 [cs]* Published Online First: 10 August 2021.http://arxiv.org/abs/2108.04949 (accessed 15 Sep 2021).
35  Yu Z, Yang X, Guo Y, *et al.* Assessing the Documentation of Social Determinants of Health for Lung Cancer Patients in Clinical Narratives. *Front Public Health* 2022;**10**:778463.
36  Guo J, Wu Y, Guo Y, *et al.* Abstract P108: Natural Language Processing Extracted Social And Behavioral Determinants Of Health And Newer Glucose-lowering Drug Initiation Among Real-world Patients With Type 2 Diabetes. *Circulation* 2022;**145**:AP108–AP108.
37  LeCun Y, Bengio Y, Hinton G. Deep learning. *Nature* 2015;**521**:436–44.
38  Lybarger K, Yetisgen M, Uzuner Ö. The 2022 n2c2/UW shared task on extracting social determinants of health. *Journal of the American Medical Informatics Association* 2023;:ocad012.
39  Liu Y, Ott M, Goyal N, *et al.* RoBERTa: A Robustly Optimized BERT Pretraining Approach. *arXiv:190711692 [cs]* Published Online First: 26 July 2019.http://arxiv.org/abs/1907.11692 (accessed 5 Mar 2021).
40  Lample G, Ballesteros M, Subramanian S, *et al.* Neural Architectures for Named Entity Recognition. In: *Proceedings of the 2016 Conference of the North American Chapter of the Association for Computational Linguistics: Human Language Technologies*. San Diego, California: : Association for Computational Linguistics 2016. 260–70.https://www.aclweb.org/anthology/N16-1030 (accessed 24 Sep 2020).
41  Devlin J, Chang M-W, Lee K, *et al.* Bert: Pre-training of deep bidirectional transformers for language understanding. *arXiv preprint arXiv:181004805* 2018.
42  Liu Y, Ott M, Goyal N, *et al.* RoBERTa: A Robustly Optimized BERT Pretraining Approach. *ArXiv* 2019;**abs/1907.11692**.
43  He P, Liu X, Gao J, *et al.* DeBERTa: Decoding-enhanced BERT with Disentangled Attention. 2021.http://arxiv.org/abs/2006.03654 (accessed 7 Feb 2023).
44  Beltagy I, Peters ME, Cohan A. Longformer: The Long-Document Transformer. 2020.http://arxiv.org/abs/2004.05150 (accessed 8 May 2023).
45  Yang X, Chen A, PourNejatian N, *et al.* A large language model for electronic health records. *npj Digit Med* 2022;**5**:1–9.
46  Yang X, Bian J, Hogan WR, *et al.* Clinical concept extraction using transformers. *Journal of the American Medical Informatics Association* 2020;**27**:1935–42.
47  Yang X, Bian J, Gong Y, *et al.* MADEx: A System for Detecting Medications, Adverse Drug Events, and Their Relations from Clinical Notes. *Drug Saf* Published Online First: 2 January 2019.

48 Yang X, Bian J, Fang R, *et al.* Identifying relations of medications with adverse drug events using recurrent convolutional neural networks and gradient boosting. *J Am Med Inform Assoc* 2020;**27**:65–72.
49 Henry S, Buchan K, Filannino M, *et al.* 2018 n2c2 shared task on adverse drug events and medication extraction in electronic health records. *J Am Med Inform Assoc* 2020;**27**:3–12.
50 Alpert J, Kim H (Julia), McDonnell C, *et al.* Barriers and Facilitators of Obtaining Social Determinants of Health of Patients With Cancer Through the Electronic Health Record Using Natural Language Processing Technology: Qualitative Feasibility Study With Stakeholder Interviews. *JMIR Form Res* 2022;**6**:e43059.

5/16/23 11:02:00 PM